%% file: main.tex
\documentclass[11pt,a4paper]{article}
\usepackage[hyperref]{acl2021}
\usepackage{times}
\usepackage{latexsym}

\usepackage{smile}
\usepackage{microtype}

\aclfinalcopy 
\def\aclpaperid{***} 

\newcommand{\hc}{highly compressed}

\title{Super Tickets in Pre-Trained Language Models: From Model Compression to Improving Generalization}

\author{\makecell{Chen Liang\thanks{~~Work was done at Microsoft Azure AI.}~~$^1$, Simiao Zuo$^1$, Minshuo Chen$^1$, Haoming Jiang$^1$, \\
Xiaodong Liu$^2$, Pengcheng He$^3$, Tuo Zhao$^1$, Weizhu Chen$^3$}\\
$^1$ Georgia Institute of Technology, $^2$ Microsoft Research, $^3$ Microsoft Azure AI \\
\texttt{\{cliang73, simiaozuo, mchen393, jianghm, tourzhao\}@gatech.edu} \\
\texttt{\{xiaodl,penhe,wzchen\}@microsoft.com}
} 

\date{}

\begin{document}
\maketitle

\input{abstract}

\input{intro}
\input{background}
\input{method_st}
\input{method_mt}
\input{experiment}

\input{analysis}
\input{conclusion}
\bibliographystyle{acl_natbib}
\bibliography{acl2021}

\clearpage
\appendix
\input{appendix}

\end{document}

%% file: abstract.tex
\begin{abstract}
\end{abstract}

The Lottery Ticket Hypothesis suggests that an over-parametrized network consists of ``lottery tickets'', and training a certain collection of them (i.e., a subnetwork) can match the performance of the full model. In this paper, we study such a collection of tickets, which is referred to as ``winning tickets'', in extremely over-parametrized models, e.g., pre-trained language models. We observe that at certain compression ratios, the generalization performance of the winning tickets can not only match but also exceed that of the full model. In particular, we observe a phase transition phenomenon: As the compression ratio increases, generalization performance of the winning tickets first improves then deteriorates after a certain threshold. We refer to the tickets on the threshold as ``super tickets''. We further show that the phase transition is task and model dependent --- as the model size becomes larger and the training data set becomes smaller, the transition becomes more pronounced. Our experiments on the GLUE benchmark show that the super tickets improve single task fine-tuning by $0.9$ points on BERT-base and $1.0$ points on BERT-large, in terms of task-average score. We also demonstrate that adaptively sharing the super tickets across tasks benefits multi-task learning\footnote{Our codes are available at \url{https://github.com/cliang1453/super-structured-lottery-tickets}.}.

%% file: intro.tex
\section{Introduction}
\label{sec:intro}

The Lottery Ticket Hypothesis (LTH, \citet{frankle2018lottery}) suggests that an over-parameterized network consists of ``lottery tickets'', and training a certain collection of them (i.e., a subnetwork) can 1) match the performance of the full model; and 2) outperform randomly sampled subnetworks of the same size (i.e., ``random tickets''). The existence of such a collection of tickets, which is usually referred to as ``winning tickets'', indicates the potential of training a smaller network to achieve the full model's performance. LTH has been widely explored in across various fields of deep learning \citep{frankle2019stabilizing,zhou2019deconstructing,you2019drawing,brix2020successfully,movva2020dissecting, girish2020lottery}. 

\begin{figure*}[!t]
    \centering
    \includegraphics[width=1.0\textwidth]{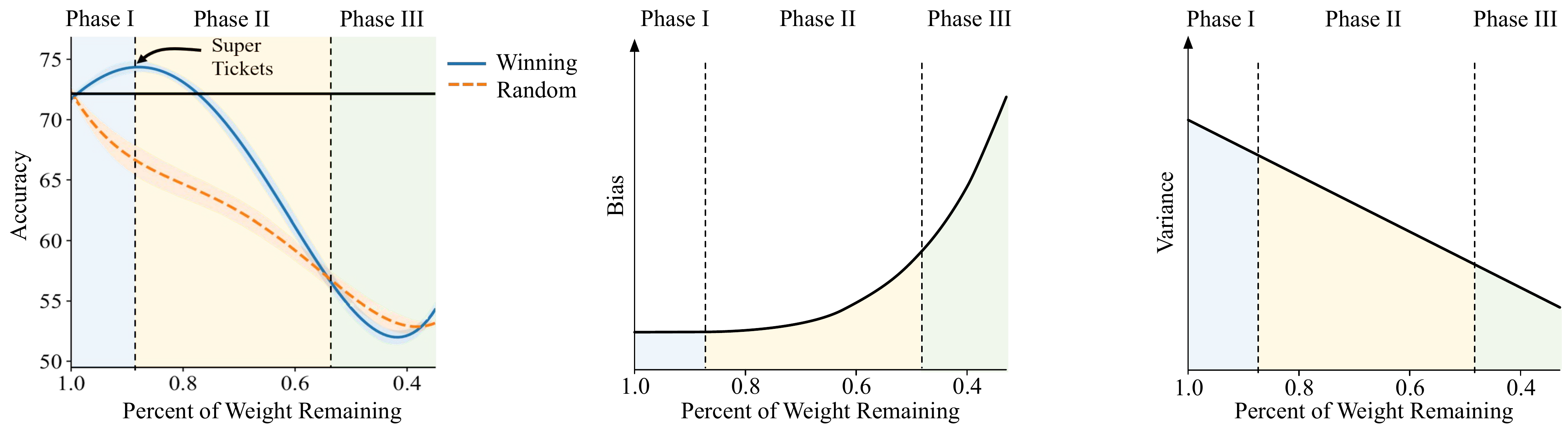}
	\caption{Illustrations of the phase transition phenomenon. \textit{Left}: Generalization performance of the fine-tuned subnetworks (the same as Figure~\ref{exp:phase_transtion} in Section~\ref{sec:st_exp}). \textit{Middle and Right}: An interpretation of bias-variance trade-off.}
	\label{fig:phase_illustration}
\end{figure*}

Aside from training from scratch, such winning tickets have demonstrated their abilities to transfer across tasks and datasets \citep{morcos2019one, yu2019playing, desai2019evaluating, chen2020lotteryb}. In natural language processing, \citet{chen2020lotterya, prasanna2020bert} have shown existence of the winning tickets in pre-trained language models. These tickets can be identified when fine-tuning the pre-trained models on downstream tasks. As the pre-trained models are usually extremely over-parameterized (e.g., BERT~\citet{devlin2018bert}, GPT-3~\citet{brown2020language}, T5~\citet{raffel2019exploring}), previous works mainly focus on searching for a highly compressed subnetwork that matches the performance of the full model. However, behavior of the winning tickets in lightly compressed subnetworks is largely overlooked.

In this paper, we study the behavior of the winning tickets in pre-trained language models, with a particular focus on lightly compressed subnetworks. We observe that generalization performance of the winning tickets selected at appropriate compression ratios can not only match, but also exceed that of the full model. In particular, we observe a \textit{phase transition} phenomenon (Figure~\ref{fig:phase_illustration}): The test accuracy improves as the compression ratio grows until a certain threshold (Phase I); Passing the threshold, the accuracy deteriorates, yet is still better than that of the random tickets (Phase II). In Phase III, where the model is \hc, training collapses. We refer to the set of winning tickets selected on that threshold as ``super tickets''.

We interpret the phase transition in the context of trade-offs between model bias and variance \citep[Chapter 7]{friedman2001elements}. 
It is well understood that an expressive model induces a small bias, and a large model induces a large variance. We classify the tickets into three categories: non-expressive tickets, lightly expressive tickets, and highly expressive tickets.
The full model has a strong expressive power due to over-parameterization, so that its bias is small. Yet its variance is relatively large.
In Phase I, by removing non-expressive tickets, variance of the selected subnetwork reduces, while model bias remains unchanged and the expressive power sustains. Accordingly, generalization performance improves.
We enter Phase II by further increasing the compression ratio. Here lightly expressive tickets are pruned. Consequently, model variance continues to decrease. However, model bias increases and overturns the benefit of the reduced variance.
Lastly for Phase III, in the \hc~region, model bias becomes notoriously large and reduction of the variance pales. As a result, training breaks down and generalization performance drops significantly.

We conduct systematic experiments and analyses to understand the phase transition. Our experiments on multiple natural language understanding (NLU) tasks in the GLUE \citep{wang2018glue} benchmark show that the super tickets can be used to improve single task fine-tuning by $0.9$ points over BERT-base \citep{devlin2018bert} and $1.0$ points over BERT-large, in terms of task-average score.
Moreover, our experiments show that the phase transition phenomenon is task and model dependent. It becomes more pronounced as a larger model is used to fit a task with less training data. In such a case, the set of super tickets forms a compressed network that exhibits a large performance gain.

The existence of super tickets suggests potential benefits to applications, such as Multi-task Learning (MTL). In MTL, different tasks require different capacities to achieve a balance between model bias and variance. However, existing methods do not specifically balance the bias and variance to accommodate each task. In fact, the fine-tuning performance on tasks with a small dataset is very sensitive to randomness. This suggests that model variance in these tasks are high due to over-parameterization. To reduce such variance, we propose a \textit{tickets sharing} strategy. Specifically, for each task, we select a set of super tickets during single task fine-tuning. Then, we adaptively share these super tickets across tasks. 

Our experiments show that tickets sharing improves MTL by $0.9$ points over MT-DNN$\textsubscript{BASE}$ \citep{liu2019multi} and $1.0$ points over MT-DNN$\textsubscript{LARGE}$, in terms of task-average score. Tickets sharing further benefits downstream fine-tuning of the multi-task model, and achieves a gain of $1.0$ task-average score. In addition, the multi-task model obtained by such a sharing strategy exhibits lower sensitivity to randomness in downstream fine-tuning tasks, suggesting a reduction in variance.

We summarize our contributions as follows:

$\bullet$ Our result is the first to identify the phase transition phenomenon in pruning large neural language models. 

$\bullet$ Our result is the first to show that pruning can improve the generalization when the models are lightly compressed, which has been overlooked by previous works. Our analysis paves the way for understanding the connection between model compression and generalization. 

$\bullet$ Motivated by our observed phase transition, we further propose a new pruning approach for multi-task fine-tuning of neural language models.

%% file: background.tex
\section{Background}
\label{sec:related}
We briefly introduce the Transformer architecture and the Lottery Ticket Hypothesis.

\subsection{Transformer Architecture}
The Transformer \citep{vaswani2017attention} encoder is composed of a stack of identical Transformer layers. Each layer consists of a multi-head attention module (MHA) followed by a feed-forward module (FFN), with a residual connection around each. The vanilla single-head attention operates as
\begin{align*}
    \text{Att}(Q, K, V) = \text{Softmax}\left(\frac{QK^{\top}}{\sqrt{d}}\right)V, 
\end{align*}
where $Q, K, V \in \RR^{l \times d}$ are $d$-dimensional vector representations of $l$ words in sequences of queries, keys and values. In MHA, the $h$-th attention head is parameterized by $W_h^Q, W_h^K, W_h^V \in \RR^{d \times d_h}$ as
\begin{align*}
    \text{H}_h (\boldsymbol{q}, \boldsymbol{x}, W_h^{\{Q,K,V\}}) = \text{Att}(\boldsymbol{q}W_h^Q,\boldsymbol{x}W_h^K,\boldsymbol{x}W_h^V),
\end{align*}
where $\boldsymbol{q}\in \RR^{l \times d}$ and $\boldsymbol{x}\in \RR^{l \times d}$ are the query and key/value vectors. In MHA, $H$ independently parameterized attention heads are applied in parallel, and the outputs are aggregated by $W_h^{O}\in \RR^{d_h \times d}$:
\begin{align*}
    \text{MHA}(\boldsymbol{q}, \boldsymbol{x})\hspace{-0.05in} = \hspace{-0.06in} \sum_h^{H} \text{H}_h (\boldsymbol{q}, \boldsymbol{x}, W_h^{\{Q,K,V\}}) W_h^{O}.
\end{align*}
Each FFN module contains a two-layer fully connected network. Given the input embedding $\boldsymbol{z}$, we let FFN($\boldsymbol{z}$) denote the output of a FFN module. 

\subsection{Structured and Unstructured LTHs}

LTH \citep{frankle2018lottery} has been widely explored in various applications of deep learning \citep{brix2020successfully,movva2020dissecting, girish2020lottery}. Most of existing results focus on finding unstructured winning tickets via iterative magnitude pruning and rewinding in randomly initialized networks \citep{frankle2019stabilizing, renda2020comparing}, where each ticket is a single parameter. Recent works further investigate learning dynamics of the tickets \citep{zhou2019deconstructing, frankle2020early} and efficient methods to identify them \citep{you2019drawing,savarese2019winning}. Besides training from scratch, researchers also explore the existence of winning tickets under transfer learning regimes for over-parametrized pre-trained models across various tasks and datasets \citep{morcos2019one, yu2019playing, desai2019evaluating, chen2020lotteryb}. For example, \citet{chen2020lotterya, prasanna2020bert} have shown the existence of winning tickets when fine-tuning BERT on downstream tasks.

There is also a surge of research exploring whether certain structures, e.g., channels in convolutional layers and attention heads in Transformers, exhibit properties of the lottery tickets. Compared to unstructured tickets, training with structured tickets is memory efficient \citep{cao2019efficient}. \citet{liu2018rethinking, prasanna2020bert} suggest that there is no clear evidence that structured winning tickets exist in randomly initialized or pre-trained weights. \citet{prasanna2020bert} observe that, in highly compressed BERT (e.g., the percent of weight remaining is around $50\%$), all tickets perform equally well. However, \citet{prasanna2020bert} have not investigated the cases where the percent of weight remaining is over $50\%$.

%% file: method_st.tex
\section{Finding Super Tickets}
\label{sec:method_st}

We identify winning tickets in BERT through structured pruning of attention heads and feed-forward layers. Specifically, in each Transformer layer, we associate mask variables $\xi_h$ to each attention head and $\nu$ to the FFN~\cite{prasanna2020bert}:
\begin{gather*}
    \text{MHA}(Q, \boldsymbol{x}) = \sum_h^{H} \xi_h \text{H}_h (Q, \boldsymbol{x}, W_h^{\{Q,K,V\}}) W_h^{O}, \\
    \text{FFN}(\boldsymbol{z}) = \nu \text{FFN}(\boldsymbol{z}).
\end{gather*}
Here, we set $\xi_h, \nu \in \{0,1\}$, and a $0$ value indicates that the corresponding structure is pruned.

We adopt importance score \citep{michel2019sixteen} as a gauge for pruning. In particular, the importance score is defined as the expected sensitivity of the model outputs with respect to the mask variables. Specifically, in each Transformer layer, 
\begin{gather*}
    I_{\text{MHA}}^h = \mathop{\EE}_{\boldsymbol{x}\sim \mathcal{D}_x}\left|\frac{\partial \mathcal{L}(\boldsymbol{x})}{\partial \xi_h}\right|, \\
    I_{\text{FFN}} = \mathop{\EE}_{\boldsymbol{x}\sim \mathcal{D}_x}\left|\frac{\partial \mathcal{L}(\boldsymbol{x})}{\partial \nu}\right|,
\end{gather*}
where $\mathcal{L}$ is a loss function and $\mathcal{D}_x$ is the data distribution. In practice, we compute the average over the training set. We apply a layer-wise $\ell_2$ normalization on the importance scores of the attention heads \citep{molchanov2016pruning, michel2019sixteen}.

The importance score is closely tied to expressive power. A low importance score indicates that the corresponding structure only has a small contribution towards the output. Such a structure has low expressive power. On the contrary, a large importance score implies high expressive power.

We compute the importance scores for all the mask variables in a single backward pass at the end of fine-tuning. We perform one-shot pruning of the same percent of heads and feed-forward layers with the lowest importance scores. We conduct pruning multiple times to obtain subnetworks, or winning tickets, at different compression ratios.

We adopt the weight rewinding technique in \citet{renda2020comparing}: We reset the parameters of the winning tickets to their values in the pre-trained weights, and subsequently fine-tune the subnetwork with the original learning rate schedule. The super tickets are selected as the winning tickets with the best rewinding validation performance. 

%% file: method_mt.tex
\section{Multi-task Learning with Tickets Sharing}
\label{sec:method_mt}

In multi-task learning, the shared model is highly over-parameterized to ensure a sufficient capacity for fitting individual tasks. Thus, the multi-task model inevitably exhibits task-dependent redundancy when being adapted to individual tasks. Such redundancy induces a large model variance.


We propose to mitigate the aforementioned model redundancy by identifying task-specific super tickets to accommodate each task's need. Specifically, when viewing an individual task in isolation, the super tickets can tailor the multi-task model to strike an appealing balance between the model bias and variance (recall from Section \ref{sec:method_st} that super tickets retain sufficient expressive power, yet keep the model variance low). Therefore, we expect that deploying super tickets can effectively tame the model redundancy for individual tasks.



\begin{figure}[!htb]
    \centering
    \includegraphics[width=0.9\linewidth]{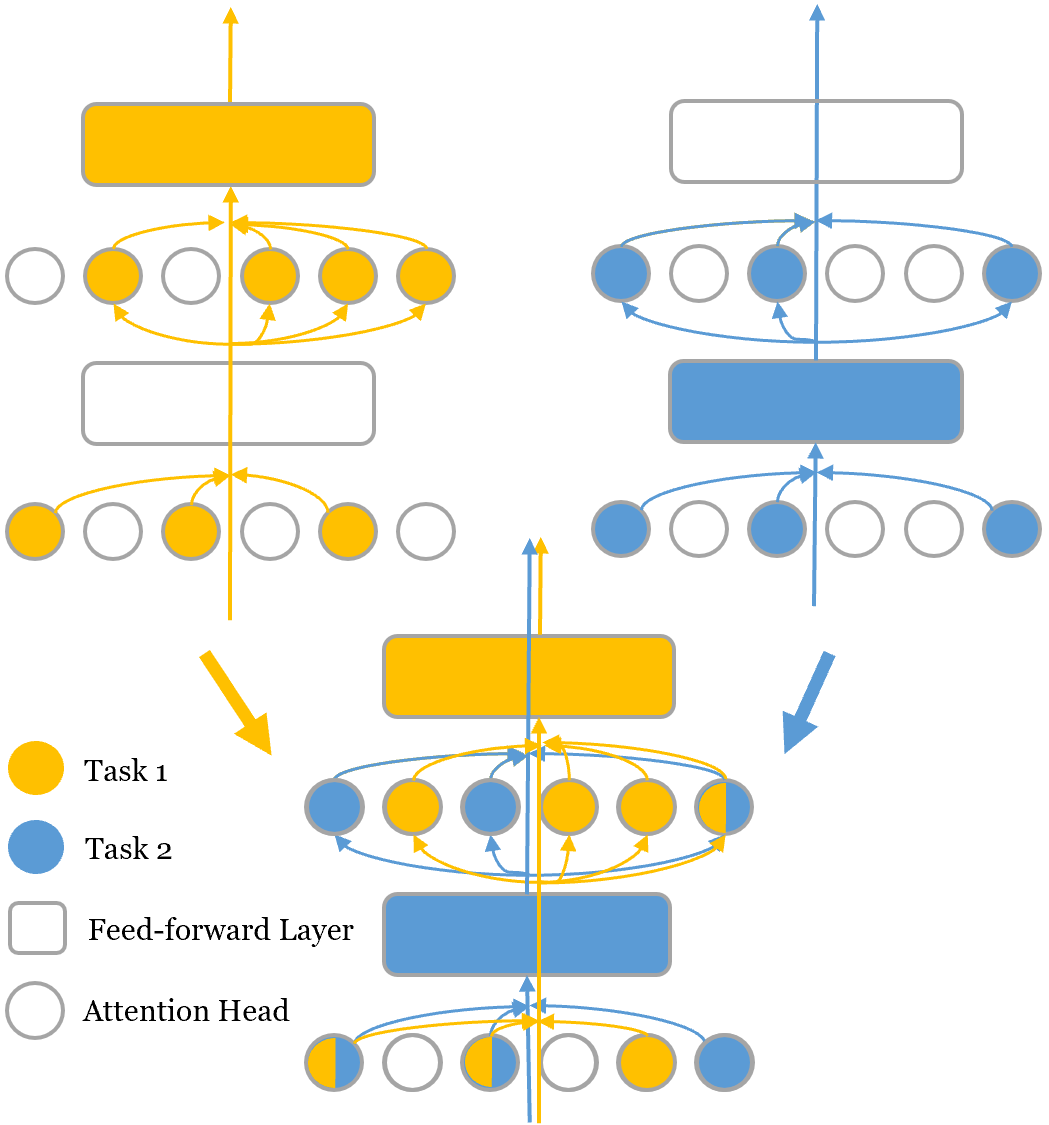}
	\caption{Illustration of tickets sharing.}
	\label{fig:sparse_sharing}
\end{figure}

Given the super tickets identified by each task, we exploit the multi-task information to reinforce fine-tuning. Specifically, we propose a \textit{tickets sharing} algorithm to update the parameters of the multi-task model: For a certain network structure (e.g., an attention head), if it is identified as super tickets by multiple tasks, then its weights are jointly updated by these tasks; if it is only selected by one specific task, then its weights are updated by that task only; otherwise, its weights are completely pruned. See Figure~\ref{fig:sparse_sharing} for an illustration.

In more detail, we denote the weight parameters in the multi-task model as $\theta$. Suppose there are $N$ tasks. For each task $i \in \{1, \dots, N\}$, we denote $\Omega^i = \{\xi^i_{h, \ell}\}_{h=1, \ell=1}^{H, L} \bigcup \{\nu^i_{\ell}\}_{\ell=1}^{L}$ as the collection of the mask variables, where $\ell$ is the layer index and $h$ is the head index. Then the parameters to be updated in task $i$ are denoted as $\theta^i = M(\theta, \Omega^i)$, where $M(\cdot, \Omega^i)$ masks the pruned parameters according to $\Omega^i$. We use stochastic gradient descent-type algorithms to update $\theta^i$. Note that the task-shared and task-specific parameters are encoded by the mask variable $\Omega^i$. The detailed algorithm is given in Algorithm \ref{algo:main}.


Tickets sharing has two major difference compared to \textit{Sparse Sharing} \citep{sun2020learning}: 1) \citet{sun2020learning} share winning tickets, while our strategy focuses on super tickets, which can better generalize and strike a sensible balance between model bias and variance. 2) In tickets sharing, tickets are structured and chosen from pre-trained weight parameters. It does not require {\it Multi-task Warmup}, which is indispensable in \citet{sun2020learning} to stabilize the sharing among unstructured tickets selected from randomly initialized weight parameters. 

\begin{algorithm}[!htb]
	\caption{Tickets Sharing}
	\label{algo:main}
	\begin{algorithmic}[1]
		\INPUT Pre-trained base model parameters $\theta$. Number of tasks $N$. Mask variables $\{\Omega^i\}_{i=1}^{N}$. Loss functions $\{\mathcal{L}^i\}_{i=1}^{N}$. Dataset $D = \bigcup_{i=1}^N D_i$. Number of epochs $T_{\max}$. 
		\For{ $i$ in $N$}
		\State Initialize the super tickets for task $i$: $\theta^i = M(\theta, \Omega^i)$. 
		\EndFor
		\For{epoch in $1,\dots,T_{\max}$}
		\State Shuffle dataset $D$.
    		\For{ a minibatch $b_i$ of task $i$ in $D$}
    		\State Compute Loss $\mathcal{L}^i(\theta^i)$. 
    		\State Compute gradient $\nabla_\theta \mathcal{L}^i(\theta^i)$.
    		\State Update $\theta^i$ using SGD-type algorithm.
		\EndFor
		\EndFor
	\end{algorithmic}
\end{algorithm}

%% file: experiment.tex
\section{Single Task Experiments}
\label{sec:st_exp}


\subsection{Data}
General Language Understanding Evaluation (GLUE,~\citet{wang2018glue}) is a standard benchmark for evaluating model generalization performance. It contains nine NLU tasks, including question answering, sentiment analysis, text similarity and textual entailment. Details about the benchmark are deferred to Appendix~\ref{app:st-data}.

\newcolumntype{C}{@{\hskip3pt}c@{\hskip3pt}}
\begin{table*}[htb!]
\centering
\small
\begin{tabular}{@{\hskip3pt}l@{\hskip3pt}|C|C|C|C|C|C|C|C|C|C}
\toprule
& \textbf{RTE} & \textbf{MRPC} & \textbf{CoLA} & \textbf{SST} & \textbf{STS-B} & \textbf{QNLI} & \textbf{QQP} & \textbf{MNLI-m/mm} & \textbf{Average} & \textbf{Average} \\
& Acc & Acc/F1 & Mcc & Acc & P/S Corr & Acc & Acc/F1 & Acc & Score & Compression \\ \midrule
ST-DNN\textsubscript{BASE}   & 69.2 & 86.2/90.4 & 57.8 & 92.9 & 89.7/89.2 & 91.2 & 90.9/88.0 & \textbf{84.5}/84.4 & 82.8 & 100\%\\ \midrule 
SuperT\textsubscript{BASE}   & \textbf{72.5} & \textbf{87.5/91.1} & \textbf{58.8} & \textbf{93.4} & \textbf{89.8/89.4} & \textbf{91.3} & \textbf{91.3/88.3} & \textbf{84.5/84.5} & \textbf{83.7} & 86.8\%\\ \midrule \midrule
ST-DNN\textsubscript{LARGE}  & 72.1 & 85.2/89.5 & 62.1 & 93.3 & 89.9/89.6 & 92.2 & 91.3/88.4 & 86.2/86.1 & 84.1 & 100\%\\ \midrule 
SuperT\textsubscript{LARGE}   & \textbf{74.1} & \textbf{88.0/91.4} & \textbf{64.3} & \textbf{93.9} & \textbf{89.9/89.7} & \textbf{92.4} & \textbf{91.4/88.5} & \textbf{86.5/86.2} & \textbf{85.1} & 81.7\%\\
\bottomrule
\end{tabular}
\caption{Single task fine-tuning evaluation results on the GLUE development set. \textit{ST-DNN} and \textit{SuperT} results are the averaged score over $5$ trails with different random seeds.}
\label{tb:st-results}
\end{table*}

\newcolumntype{C}{@{\hskip3pt}c@{\hskip3pt}}
\begin{table*}[htb!]
\centering
\small
\begin{tabular}{@{\hskip3pt}l@{\hskip3pt}|C|C|C|C|C|C|C|C|C|C}
\toprule
& \textbf{RTE} & \textbf{MRPC} & \textbf{CoLA} & \textbf{SST} & \textbf{STS-B} & \textbf{QNLI} & \textbf{QQP} & \textbf{MNLI-m/mm} & \textbf{Average} & \textbf{Average} \\
& Acc & F1 & Mcc & Acc & S Corr & Acc & F1 & Acc & Score & Compression \\ \midrule
ST-DNN\textsubscript{BASE}    & 66.4 & 88.9   & 52.1 & 93.5 & 85.8    & \textbf{90.5} &  71.2   & \textbf{84.6}/83.4 & 79.6 & 100\%\\ \midrule 
SuperT\textsubscript{BASE}    & \textbf{69.6} & \textbf{89.4} & \textbf{54.3} & \textbf{94.1} & \textbf{86.2} & \textbf{90.5} & \textbf{71.3} & \textbf{84.6/83.8} & \textbf{80.4} & 86.8\%\\ 
\bottomrule
\end{tabular}
\caption{Single task fine-tuning test set results scored using the GLUE evaluation server\footnote{\url{https://gluebenchmark.com/leaderboard}}. Results of \textit{ST-DNN\textsubscript{BASE}} are from \citet{devlin2018bert}.}
\label{tb:st-results-test}
\end{table*}

\subsection{Models \& Training}
\label{sec:st_exp_training}
We fine-tune a pre-trained BERT model with task-specific data to obtain a single task model. We append a task-specific fully-connected layer to BERT as in \citet{devlin2018bert}. 

\noindent $\bullet$ $\textbf{ST-DNN\textsubscript{BASE/LARGE}}$ is initialized with BERT-base/large followed by a task-specific layer. \\
\noindent $\bullet$ $\textbf{SuperT\textsubscript{BASE/LARGE}}$ is initialized with the chosen set of super tickets in BERT-base/large followed by a task-specific layer. Specifically, we prune BERT-base/large in unit of $10\%$ heads and $10\%$ feed-forward layers (FFN) at $8$ different sparsity levels (10\% heads and 10\% FFN, 20\% heads and 20\% FFN, etc). Among them, the one with the best rewinding validation result is chosen as the set of super tickets. We randomly sample $10\%$ GLUE development set for tickets selection. 

Our implementation is based on the MT-DNN code base\footnote{https://github.com/namisan/mt-dnn}. We use Adamax \citep{kingma2014adam} as our optimizer. We tune the learning rate in $\{5\times10^{-5}, 1\times10^{-4}, 2\times 10^{-4}\}$ and batch size in $\{8, 16, 32\}$. We train for a maximum of $6$ epochs with early-stopping. All training details are summarized in Appendix~\ref{app:st-training}. 

\subsection{Generalization of the Super Tickets}
We conduct $5$ trails of pruning and rewinding experiments using different random seeds. Table~\ref{tb:st-results} and~\ref{tb:st-results-test} show the averaged evaluation results on the GLUE development and test sets, respectively. We remark that the gain of SuperT$\textsubscript{BASE/LARGE}$ over ST-DNN$\textsubscript{BASE/LARGE}$ is statistically significant. All the results\footnote{Except for STS-B (SuperT\textsubscript{BASE},  Table~\ref{tb:st-results}), where the p-value is $0.37$.} have passed a paired student t-test with p-values less than $0.05$. More validation statistics are summarized in Appendix~\ref{app:st-val-stats}.

Our results can be summarized as follows.

1) In all the tasks, SuperT consistently achieves better generalization than ST-DNN. The task-averaged improvement is around $0.9$ over ST-DNN$\textsubscript{BASE}$ and $1.0$ over ST-DNN$\textsubscript{LARGE}$.

\begin{figure}[!htb]
    \centering
    \includegraphics[width=1.0\linewidth]{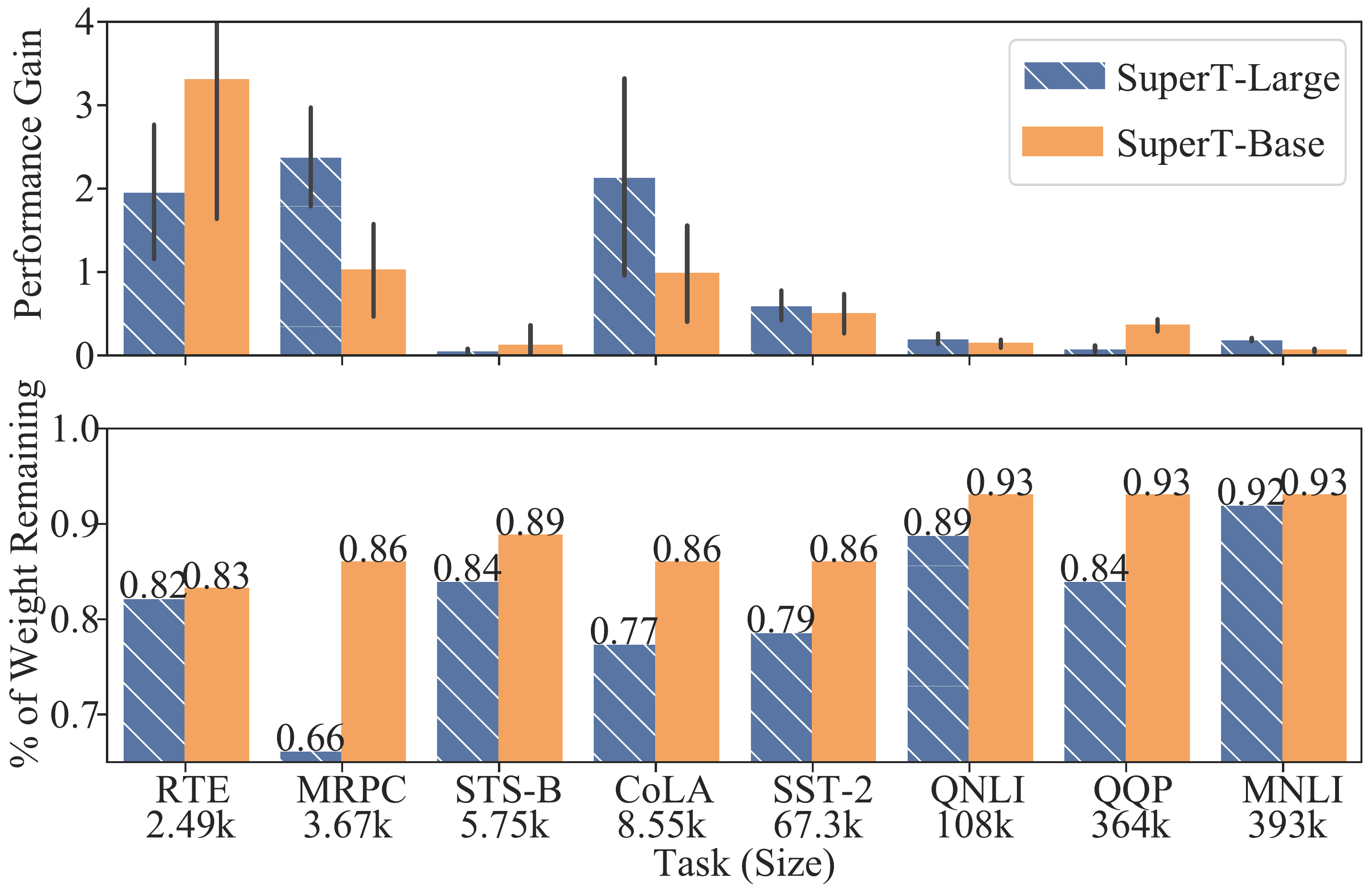}
	\caption{Single task fine-tuning validation results in different GLUE tasks. \textit{Upper}: Performance Gain. \textit{Lower}: Percent of weight remaining.}
	\label{exp:st-results}
\end{figure}

2) Performance gain of the super tickets is more significant in small tasks. For example, in Table~\ref{tb:st-results}, we obtain $3.3$ points gain on RTE (2.5k data), but only $0.4/0.3$ on QQP (364k data) in the SuperT\textsubscript{BASE} experiments.
Furthermore, from Figure~\ref{exp:st-results}, note that the super tickets are more heavily compressed in small tasks, e.g., for SuperT\textsubscript{BASE}, $83\%$ weights remaining for RTE, but $93\%$ for QQP.
These observations suggest that for small tasks, model variance is large, and removing non-expressive tickets reduces variance and improves generalization. For large tasks, model variance is low, and all tickets are expressive to some extent.

3) Performance of the super tickets is related to model size. Switching from SuperT$\textsubscript{BASE}$ to SuperT$\textsubscript{LARGE}$, the percent of weights remaining shrinks uniformly across tasks, yet the generalization gains persist (Figure~\ref{exp:st-results}). This suggests that in large models, more non-expressive tickets can be pruned without performance degradation.

\begin{figure}[!htb]
    \centering
    \includegraphics[width=0.98\linewidth]{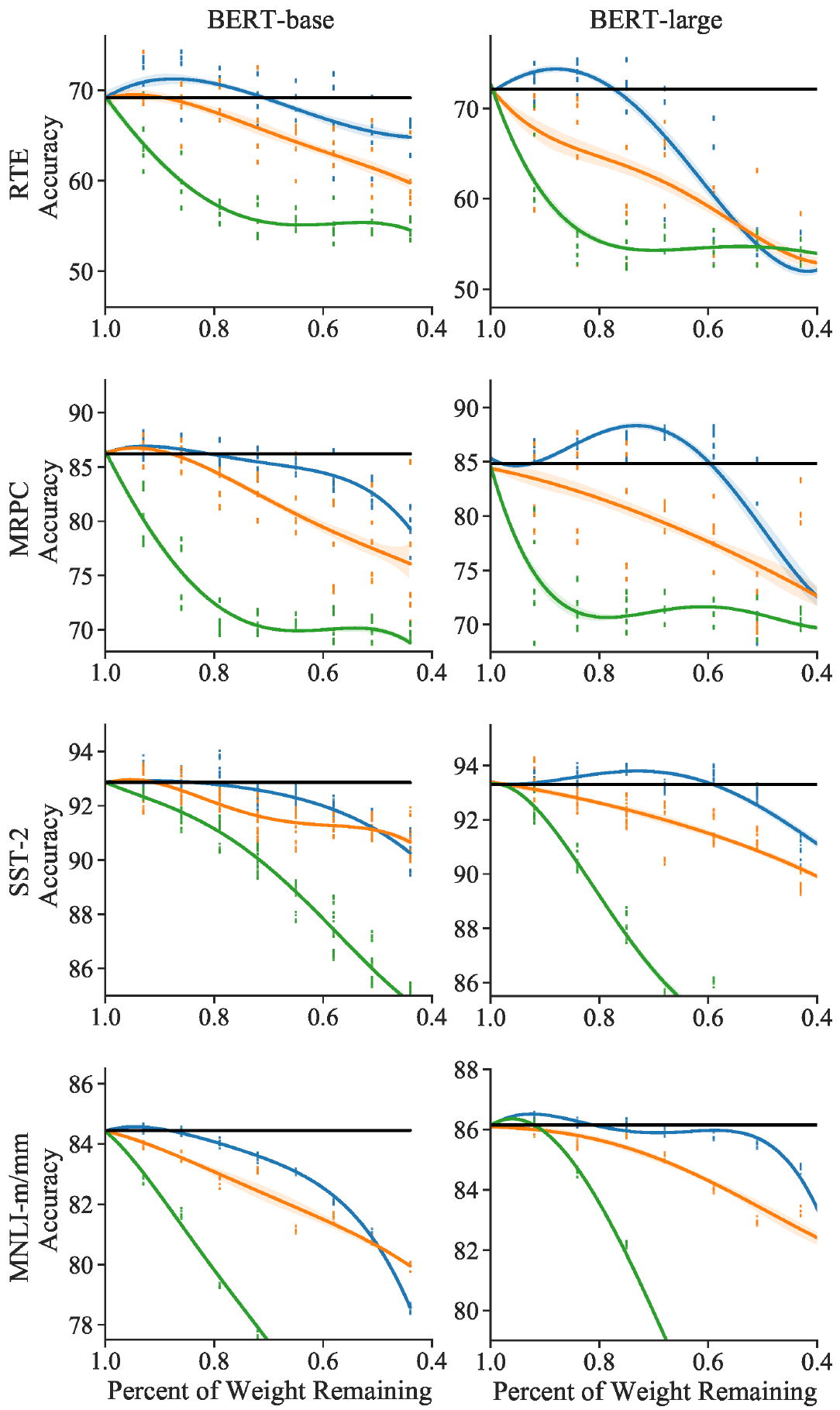}
	\caption{Single task fine-tuning evaluation results of the winning (blue), the random (orange), and the losing (green) tickets on the GLUE development set under various sparsity levels.}
	\label{exp:phase_transtion}
\end{figure}
\begin{figure}[!htb]
    \includegraphics[width=0.98\linewidth, height=0.44\linewidth]{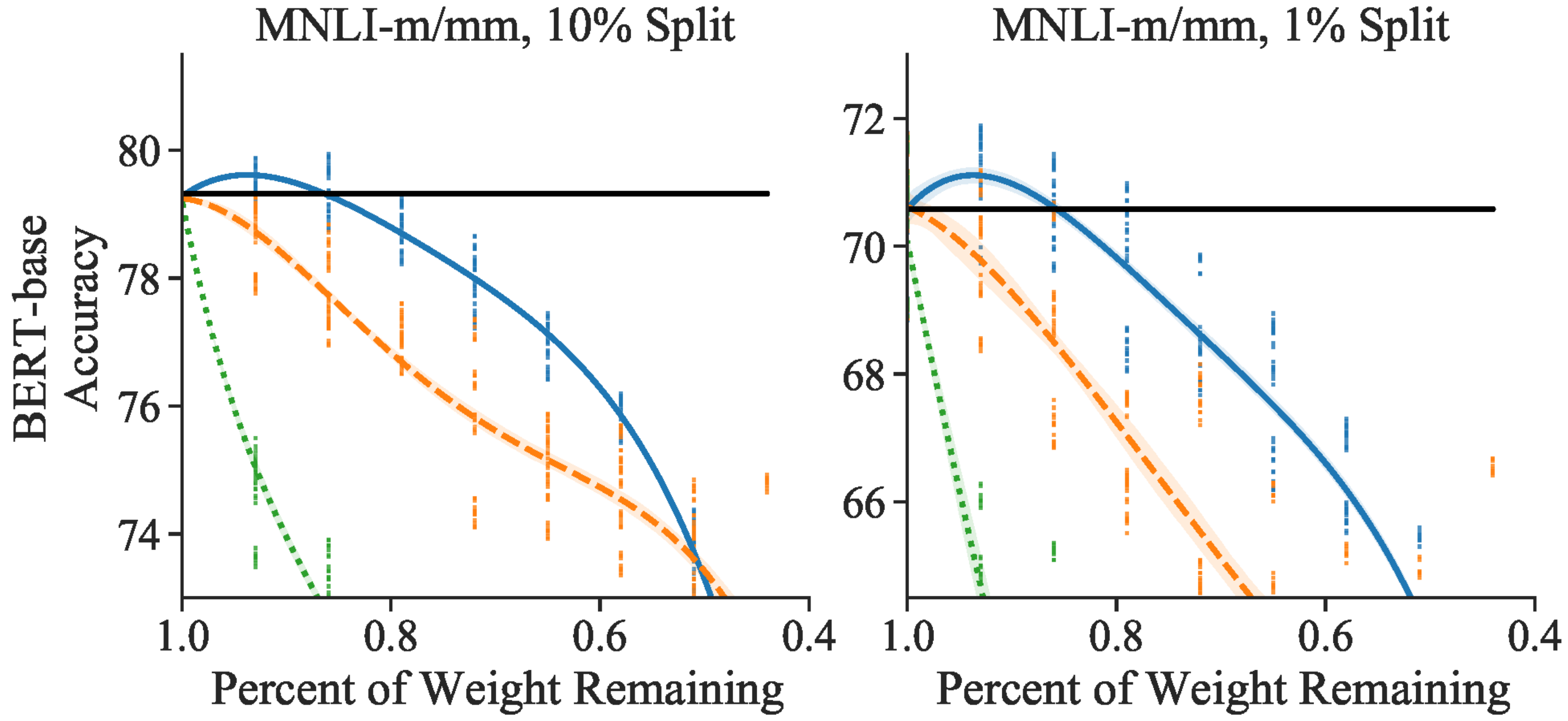}
    \caption{Phase transition under different randomly sampled training subsets. Note that the settings are the same as Figure~\ref{exp:phase_transtion} (bottom left), except the data size.}
    \label{exp:phase_split}
\end{figure}

\subsection{Phase Transition}

Phase transitions are shown in Figure~\ref{exp:phase_transtion}. We plot the evaluation results of the winning, the random, and the losing tickets under $8$ sparsity levels using BERT-base and BERT-large.
The winning tickets contain structures with the highest importance scores. The losing tickets are selected reversely, i.e., the structures with the lowest importance scores are selected, and high-importance structures are pruned. The random tickets are sampled uniformly across the network. We plot the averaged scores over $5$ trails using different random seeds\footnote{Except for MNLI, where we plot $3$ trails as the there are less variance among trails.}. Phase transitions of all the GLUE tasks are in Appendix~\ref{app:phase-transition}.

We summarize our observations:

\begin{table*}[htb!]
\centering
\small
\begin{tabular}{@{\hskip3pt}l@{\hskip3pt}|C|C|C|C|C|C|C|C|C|C} \toprule
& \textbf{RTE} & \textbf{MRPC} & \textbf{CoLA} & \textbf{SST} & \textbf{STS-B} & \textbf{QNLI} & \textbf{QQP} & \textbf{MNLI-m/mm} & \textbf{Average} & \textbf{Average} \\
& Acc & Acc/F1 & Mcc & Acc & P/S Corr & Acc & Acc/F1 & Acc & Score & Compression \\ \midrule
MT-DNN\textsubscript{BASE}    & 79.0 & 80.6/86.2 & 54.0 & 92.2 & 86.2/86.4 & 90.5 & 90.6/87.4 & 84.6/84.2 & 82.4 & 100\%\\
 + ST Fine-tuning   & 79.1 & 86.8/89.2 & 59.5 & \textbf{93.6} & 90.6/90.4 & 91.0 & \textbf{91.6}/88.6 & \textbf{85.3/85.0} & 84.6 & 100\%\\ \midrule 
Ticket-Share\textsubscript{BASE}   & 81.2 & 87.0/90.5 & 52.0 & 92.7 & 87.7/87.5 & 91.0 & 90.7/87.5 & 84.5/84.1 & 83.3 & 92.9\%\\ 
 + ST Fine-tuning   & \textbf{83.0} & \textbf{89.2/91.6} & \textbf{59.7} & 93.5 & \textbf{91.1/91.0} & \textbf{91.9}  & \textbf{91.6/88.7}  & 85.0/\textbf{85.0}  &  \textbf{85.6} & 92.9\%\\ \midrule \midrule 
MT-DNN\textsubscript{LARGE}   & 83.0 & 85.2/89.4 & 56.2 & 93.5 & 87.2/86.9 & 92.2 & 91.2/88.1 & 86.5/86.0 & 84.4 & 100\%\\ 
 + ST Fine-tuning   & 83.4 & 87.5/91.0 & 63.5 & \textbf{94.3} & 90.7/90.6 & 92.9 & \textbf{91.9/89.2} & \textbf{87.1}/86.7 & 86.4 & 100\% \\ \midrule 
Ticket-Share\textsubscript{LARGE}    & 80.5 & 88.4/91.5 & 61.8 & 93.2 & 89.2/89.1 & 92.1 & 91.3/88.4 & 86.7/86.0 & 85.4 & 83.3\%\\ 
 + ST Fine-tuning   & \textbf{84.5} & \textbf{90.2}/\textbf{92.9} & \textbf{65.0} & 94.1 & \textbf{91.3}/\textbf{91.1} & \textbf{93.0} & \textbf{91.9}/89.1 & 87.0/\textbf{86.8} & \textbf{87.1} & 83.3\%\\ \bottomrule
\end{tabular}
\caption{Multi-task Learning evaluation results on the GLUE development set. Results of \textit{MT-DNN\textsubscript{BASE/LARGE}} with and without ST Fine-tuning are from \citet{liu2020mtmtdnn}.}
\label{tb:mt-results}
\end{table*}

1) The winning tickets are indeed the ``winners''. In Phase I and early Phase II, the winning tickets perform better than the full model and the random tickets. This demonstrates the existence of structured winning tickets in lightly compressed BERT models, which \citet{prasanna2020bert} overlook.

2) Phase transition is pronounced over different tasks and models. Accuracy of the winning tickets increases up till a certain compression ratio (Phase I); Passing the threshold, the accuracy decreases (Phase II), until its value intersects with that of the random tickets (Phase III). Note that Phase III agrees with the observations in \citet{prasanna2020bert}.
Accuracy of the random tickets decreases in each phase. This suggests that model bias increases steadily, since tickets with both low and high expressive power are discarded.
Accuracy of the losing tickets drops significantly even in Phase I, suggesting that model bias increases drastically as highly expressive tickets are pruned. 

3) Phase transition is more pronounced in large models and small tasks.
For example, in Figure~\ref{exp:phase_transtion}, the phase transition is more noticeable in BERT-large than in BERT-base, and is more pronounced in RTE (2.5k) and MRPC (3.7k) than in SST (67k) and MNLI (393k).
The phenomenon becomes more significant for the same task when we only use a part of the data, e.g., Figure~\ref{exp:phase_split} vs. Figure~\ref{exp:phase_transtion} (bottom left).

\section{Multi-task Learning Experiments}
\label{sec:mt_exp}

\subsection{Model \& Training}

We adopt the MT-DNN architecture proposed in \citet{liu2020mtmtdnn}. The MT-DNN model consists of a set of task-shared layers followed by a set of task-specific layers. The task-shared layers take in the input sequence embedding, and generate shared semantic representations by optimizing multi-task objectives. Our implementation is based on the MT-DNN code base. We follow the same training settings in \citet{liu2020mtmtdnn} for multi-task learning, and in Section~\ref{sec:st_exp_training} for downstream fine-tuning. More details are summarized in Appendix~\ref{app:mtl-training}. 

\noindent $\bullet$ $\textbf{MT-DNN\textsubscript{BASE/LARGE}.}$ An MT-DNN model refined through multi-task learning, with task-shared layers initialized by pre-trained BERT-base/large.\\
\noindent $\bullet$ $\textbf{MT-DNN\textsubscript{BASE/LARGE} + ST Fine-tuning.}$ A single task model obtained by further fine-tuning MT-DNN on an individual downstream task.\\
\noindent $\bullet$ $\textbf{Ticket-Share\textsubscript{BASE/LARGE}.}$ An MT-DNN model refined through the ticket sharing strategy, with task-shared layers initialized by the union of the super tickets in pre-trained BERT-base/large. \\
\noindent $\bullet$ $\textbf{Ticket-Share\textsubscript{BASE/LARGE} + ST Fine-tuning.}$ A fine-tuned single-task \textbf{Ticket-Share} model.

\subsection{Experimental Results}

Table~\ref{tb:mt-results} summarizes experimental results. The fine-tuning results are averaged over $5$ trails using different random seeds. We have several observations: 

1) Ticket-Share$\textsubscript{BASE}$ and Ticket-Share$\textsubscript{LARGE}$ achieve $0.9$ and $1.0$ gain in task-average score over MT-DNN$\textsubscript{BASE}$ and MT-DNN$\textsubscript{LARGE}$, respectively.
In some small tasks (RTE, MRPC), Ticket-Share achieves better or on par results compared to MT-DNN+Fine-tuning. This suggests that by balancing the bias and variance for different tasks, the multi-task model's variance is reduced.
In large tasks (QQP, QNLI and MNLI), Ticket-Share behaves equally well with the full model. This is because task-shared information is kept during pruning and still benefits multi-task learning.

2) Ticket-Share$\textsubscript{BASE}$+Fine-tuning and Ticket-Share$\textsubscript{LARGE}$+Fine-tuning achieve $1.0$ and $0.7$ gains in task-average score over MT-DNN$\textsubscript{BASE}$+Fine-tuning and MT-DNN$\textsubscript{LARGE}$+Fine-tuning, respectively. This suggests that reducing the variance in the multi-task model benefits fine-tuning downstream tasks.

\begin{table}[htb!]
\small
\begin{tabular}{lllll}
\toprule
\multicolumn{1}{l|}{Model}                              & 0.1\% & 1\%  & 10\%  & 100\%  \\ \midrule
\multicolumn{5}{c}{\textbf{SNLI} (Dev Acc\%)}                                 \\ 
\multicolumn{1}{l}{\# Training Data}                     & 549   & 5493 & 54k & 549k \\ \midrule
\multicolumn{1}{l|}{$\text{MNLI-ST-DNN}\textsubscript{BASE}$}    &  82.1  &  85.1    & 88.4      & 90.7   \\
\multicolumn{1}{l|}{$\text{MNLI-SuperT}\textsubscript{BASE}$}    &  \textbf{82.9}  &  \textbf{85.5}    & \textbf{88.8}      & \textbf{91.4}   \\\midrule
\multicolumn{1}{l|}{$\text{MT-DNN}\textsubscript{BASE}$}         & 82.1   & 85.2     & 88.4      & 91.1   \\
\multicolumn{1}{l|}{$\text{Ticket-Share}\textsubscript{BASE}$}   & \textbf{83.3}   & \textbf{85.8}     & \textbf{88.9}      & \textbf{91.5}   \\ \midrule
\multicolumn{5}{c}{\textbf{SciTail} (Dev Acc\%)}                                 \\ 
\multicolumn{1}{l}{\# Training Data}                    &  23  & 235 & 23k & 235k \\ \midrule
\multicolumn{1}{l|}{$\text{MNLI-ST-DNN}\textsubscript{BASE}$}   &  80.6 & 88.8  & 92.0   &  95.7  \\
\multicolumn{1}{l|}{$\text{MNLI-SuperT}\textsubscript{BASE}$}   &  \textbf{82.9} & \textbf{89.8}  & \textbf{92.8}   &  \textbf{96.2}      \\ \midrule
\multicolumn{1}{l|}{$\text{MT-DNN}\textsubscript{BASE}$}        & 81.9  & 88.3 & 91.1    & 95.7   \\
\multicolumn{1}{l|}{$\text{Ticket-Share}\textsubscript{BASE}$}  &  \textbf{83.1}  &  \textbf{90.1} &  \textbf{93.5}  &   \textbf{96.5}     \\ \bottomrule
\end{tabular}
\caption{Domain adaptation evaluation results on SNLI and SciTail development set. Results of $\textit{MT-DNN\textsubscript{BASE}}$ are from \citet{liu2020mtmtdnn}.}
\label{tb:ood-adapt-results}
\end{table}

\begin{figure*}[!htb]
    \centering
    \includegraphics[width=1.0\textwidth]{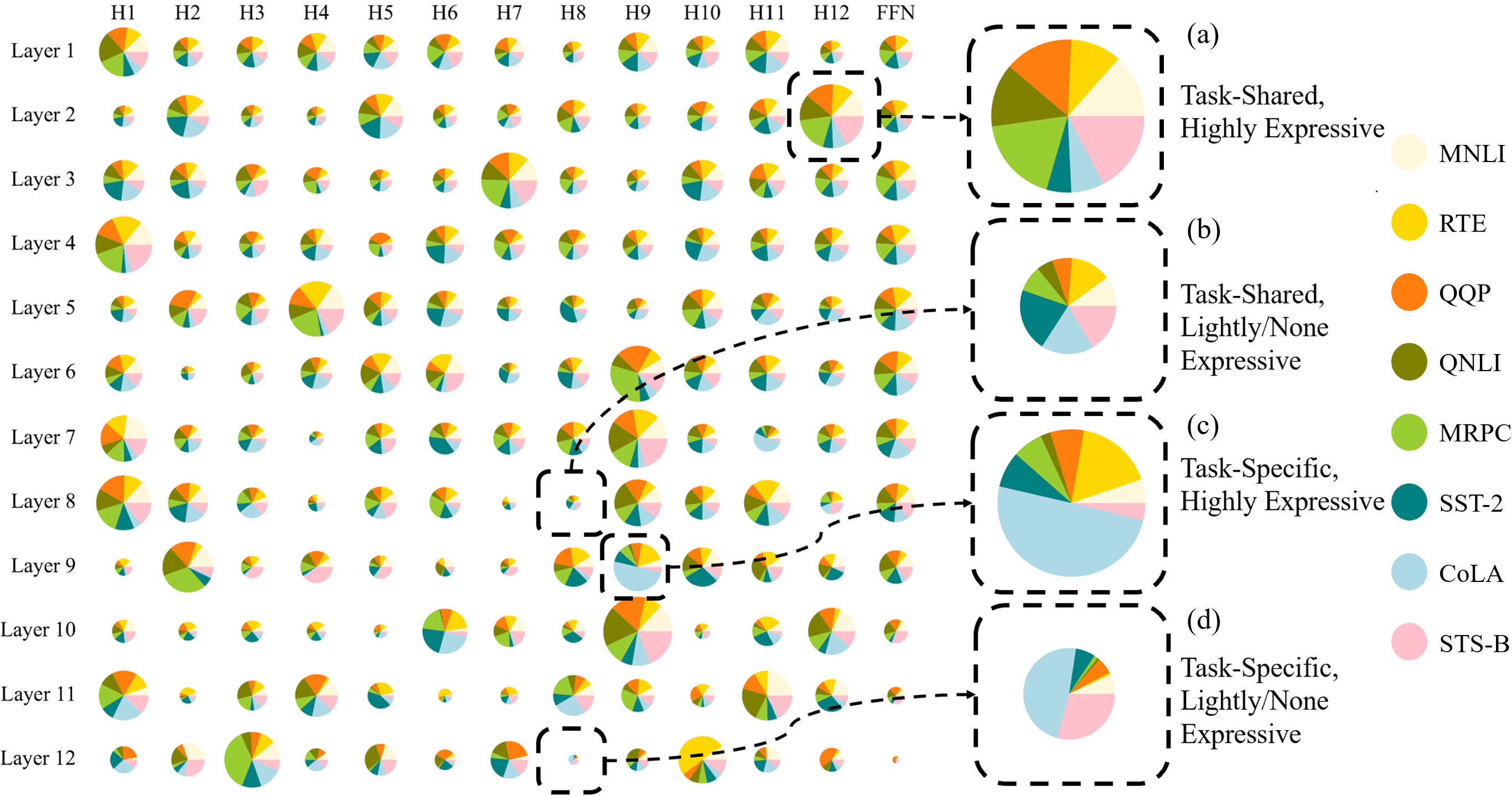}
	\caption{Illustration of tickets importance across tasks. Each ticket is represented by a pie chart. The size of a pie indicates the \textit{Ticket Importance}, where a larger pie suggests the ticket exhibits higher expressivity. Each task is represented by a color. The share of a color indicates the \textit{Task Share}, where a even share suggests the ticket exhibits equal expressivity in all tasks.}
	\label{fig:head_task_analysis}
\end{figure*}

\section{Domain Adaptation}

To demonstrate that super tickets can quickly generalize to new tasks/domains, we conduct few-shot domain adaptation on out-of-domain NLI datasets.

\subsection{Data \& Training}
We briefly introduce the target domain datasets. The data and training details are summarized in Appendix~\ref{app:da-data} and~\ref{app:da-training}, respectively.

\noindent $\textbf{SNLI}.$ The Stanford Natural Language Inference dataset \citep{bowman2015large} is one of the most widely used entailment dataset for NLI.
It contains $570$k sentence pairs, where the premises are drawn from the captions of the Flickr30 corpus and hypotheses are manually annotated.  

\noindent $\textbf{SciTail}$ is a textual entailment dataset derived from a science question answering (SciQ) dataset \citep{khot2018scitail}. The hypotheses are created from science questions, rendering SciTail challenging. 

\subsection{Experimental Results}

We consider domain adaptation on both single task and multi-task super tickets. Specifically, we adapt $\text{SuperT}_\text{BASE}$ and $\text{ST-DNN}_\text{BASE}$ from MNLI to SNLI/SciTail, and adapt the shared embeddings generated by $\text{Ticket-Share}_\text{BASE}$ and by $\text{MT-DNN}_\text{BASE}$ to SNLI/SciTail. We adapt these models to $0.1\%, 1\%, 10\%$ and $100\%$ SNLI/SciTail training sets\footnote{We use the subsets released in MT-DNN code base.}, and evaluate the transferred models on SNLI/SciTail development sets. Table~\ref{tb:ood-adapt-results} shows the domain adaptation evaluation results. As we can see, SuperT and Ticket-Share can better adapt to SNLI/SciTail than ST-DNN and MT-DNN, especially under the few shot setting.

%% file: analysis.tex
\section{Analysis}

\noindent \textbf{Sensitivity to Random Seed.} To better demonstrate that training with super tickets effectively reduces model variance, we evaluate models' sensitivity to changes in random seeds during single task fine-tuning and multi-task downstream fine-tuning. In particular, we investigate fitting small tasks with highly over-parametrized models (variance is often large in these models, see Section~\ref{sec:st_exp} and~\ref{sec:mt_exp}). As shown in Table~\ref{tb:seed_study}, SuperT$\textsubscript{LARGE}$ and Ticket-Share$\textsubscript{LARGE}$ induce much smaller standard deviation in validation results. Experimental details and further analyses are deferred to Appendix~\ref{app:ana-sensitivity}.

\begin{table}[htb!]
\centering
\small
\begin{tabular}{@{\hskip3pt}l@{\hskip3pt}|C|C|C|C|C}
\toprule
& \textbf{RTE} & \textbf{MRPC} & \textbf{CoLA} & \textbf{STS-B} & \textbf{SST-2} \\ \midrule 
ST-DNN\textsubscript{LARGE}  & 1.17 & 0.61 & 1.32 & 0.16 & 0.17 \\ 
SuperT\textsubscript{LARGE}  & 0.72 & 0.20 & 0.97 & 0.07 & 0.16 \\ \midrule
MT-DNN\textsubscript{LARGE}   & 1.43 & 0.78 & 1.14  & 0.15 & 0.18  \\ 
Ticket Share\textsubscript{LARGE}   & 0.99 & 0.67 & 0.81 & 0.08 & 0.16  \\ \bottomrule
\end{tabular}
\caption{Standard deviation of tasks in GLUE (dev) over $5$ different random seeds.}
\label{tb:seed_study}
\end{table}

\noindent \textbf{Tickets Importance Across Tasks.} We analyze the importance score of each ticket computed in different GLUE tasks. For each ticket, we compute the importance score averaged over tasks as the \textit{Ticket Importance}, and the proportion of the task-specific importance score out of the sum of all tasks' scores as the \textit{Task Share}, as illustrated in Figure~\ref{fig:head_task_analysis}. 

We observe that many tickets exhibit almost equal \textit{Task Share}s for over $5$ out of $8$ tasks (Figure~\ref{fig:head_task_analysis}(a)(b)). While these tickets contribute to the knowledge sharing in the majority of tasks, they are considered non-expressive for tasks such as SST-2 (see Figure \ref{fig:head_task_analysis}(a)(c)(d)). This explains why SST-2 benefits little from tickets sharing. Furthermore, a small number of tickets are dominated by a single task, e.g., CoLA (Figure~\ref{fig:head_task_analysis}(c)), or dominated jointly by two tasks, e.g., CoLA and STS-B (Figure~\ref{fig:head_task_analysis}(d)). This suggests that some tickets only learn task-specific knowledge, and the two tasks may share certain task-specific knowledge.

%% file: conclusion.tex

\section{Discussion}
\label{sec:discussion}

\noindent \textbf{Structured Lottery Tickets.} LTH hypothesizes that a subset of unstructured parameters can be trained to match the full model's performance. Instead, we question whether a subset of structured weight matrices, e.g., FFN layers and attention heads, can also be trained to match the full model's performance. This question is more practically important than the unstructured one: training and inference on structured matrices are better optimized for hardware acceleration. Our results give a positive answer to this question, while previous works show that the structured tickets do not exist in highly compressed models \citep{prasanna2020bert}. 

\noindent \textbf{Searching Better Generalized Super Tickets.} We select winning tickets according to the sensitivity of the model outputs with respect to the mask variables of each structure \citep{michel2019sixteen, prasanna2020bert}, as this measure is closely tied to the structure's expressive power (Section~\ref{sec:method_st}). In addition, we conduct an one-shot pruning for computational simplicity. We leave other importance measures and pruning schedules, which may help identifying better generalized super tickets, for future works \citep{voita2019analyzing, behnke-heafield-2020-losing, wang2019structured, fan2019reducing, zhou2020scheduled, sajjad2020poor}.

\noindent \textbf{Searching Super Tickets Efficiently.} Determining the compression ratio of the super tickets requires rewinding models at multiple sparsity levels. To leverage super tickets in practice, a potential direction of research is to find heuristics to determine this ratio prior or early-on in training. We leave this for future works.

\section{Conclusion}
\label{sec:conclusion}

We study the behaviors of the structured lottery tickets in pre-trained BERT. We observe that the generalization performance of the winning tickets exhibits a phase transition phenomenon, suggesting pruning can improve generalization when models are lightly compressed. Based on the observation, we further propose a tickets sharing strategy to improve multi-task fine-tuning. Our analysis paves the way for understanding the connection between model compression and generalization. 

\section*{Broader Impact}

This paper studies the behavior of the structured lottery tickets in pre-trained language models. Our investigation neither introduces any social/ethical bias to the model nor amplifies any bias in the data. We do not foresee any direct social consequences or ethical issues. Furthermore, our proposed method improves performance through model compression, rendering it energy efficient.

%% file: appendix.tex
\section{Appendix}
\label{sec:appendix}

\subsection{Single Task Experiments}
\subsubsection{Data}
\label{app:st-data}
\textbf{GLUE.} GLUE is a collection of nine NLU tasks. The benchmark includes question answering~\cite{squad1}, linguistic acceptability (CoLA, \citealt{cola2018}), sentiment analysis (SST, \citealt{sst2013}), text similarity (STS-B, \citealt{sts-b2017}), paraphrase detection (MRPC, \citealt{mrpc2005}), and natural language inference (RTE \& MNLI, \citealt{rte1,rte2,rte3,rte5,mnli2018}) tasks. Details of the GLUE benchmark, including tasks, statistics, and evaluation metrics, are summarized in Table~\ref{tab:glue}.

\subsubsection{Training}
\label{app:st-training}
We use Adamax as the optimizer. A linear learning rate decay schedule with warm-up over $0.1$ is used. We apply a gradient norm clipping of $1$. We set the dropout rate of all task specific layers as $0.1$, except $0.3$ for MNLI and $0.05$ for CoLA. All the texts were tokenized using wordpieces, and were chopped to spans no longer than $512$ tokens. All experiments are conducted on Nvidia V100 GPUs.

\subsubsection{Evaluation Results Statistics}
\label{app:st-val-stats}

We conduct $5$ sets of experiments on different random seeds. Each set of experiment consists of fine-tuning, pruning, and rewinding at $8$ sparsity levels. For results on GLUE dev set (Table~\ref{tb:st-results}), we report the average score of super tickets rewinding results over $5$ sets of experiments. The standard deviation of the results is shown in Table~\ref{tb:st-val-stats}. The statistics of the percent of weight remaining in the selected super tickets are shown in Table~\ref{tb:st-compress-stats}. 

\begin{table*}[htb!]
\centering
\small
\begin{tabular}{@{\hskip3pt}l@{\hskip3pt}|C|C|C|C|C|C|C|C} \toprule
& \textbf{RTE} & \textbf{MRPC} & \textbf{CoLA} & \textbf{STS-B} & \textbf{SST-2} & \textbf{QNLI} & \textbf{QQP} & \textbf{MNLI} \\ \midrule 
SuperT\textsubscript{BASE}  & 0.91 & 0.74 & 1.51 & 0.49 & 0.50 & 0.10 & 0.08 & 0.04\\ 
SuperT\textsubscript{LARGE} & 0.72 & 0.20 & 0.97 & 0.07 & 0.16 & 0.07 & 0.11 & 0.02 \\ \bottomrule
\end{tabular}
\caption{Standard deviation of the evaluation results on GLUE development set over $5$ different random seeds.}
\label{tb:st-val-stats}
\end{table*}

\begin{table*}[htb!]
\centering
\small
\begin{tabular}{@{\hskip3pt}l@{\hskip3pt}|C|C|C|C|C|C|C|C} \toprule
& \textbf{RTE} & \textbf{MRPC} & \textbf{CoLA} & \textbf{STS-B} & \textbf{SST-2} & \textbf{QNLI} & \textbf{QQP} & \textbf{MNLI} \\ \midrule
SuperT\textsubscript{BASE} (Mean)  & 0.83 & 0.86 & 0.89 & 0.86 & 0.93 & 0.93 & 0.93 & 0.93\\
SuperT\textsubscript{BASE} (Std Dev)  & 0.07 & 0.08 & 0.04 & 0.06 & 0.07 & 0.00 & 0.00 & 0.00\\
SuperT\textsubscript{LARGE} (Mean) & 0.82 & 0.66 & 0.84 & 0.77 & 0.79 & 0.90 & 0.84 & 0.92 \\
SuperT\textsubscript{LARGE} (Std Dev) & 0.04 & 0.04 & 0.00 & 0.10 & 0.05 & 0.03 & 0.00 & 0.00 \\
\bottomrule
\end{tabular}
\caption{Statistics of the percent of weight remaining of the selected super tickets over $5$ different random seeds.}
\label{tb:st-compress-stats}
\end{table*}

For results on GLUE test set (Table~\ref{tb:st-results-test}), as the evaluation server sets an limit on submission times, we only evaluate the test prediction under a single random seed that gives the best task-average validation results.

\subsection{Multi-task Learning Experiments}
\label{app:mtl-training}
\subsubsection{Multi-task Model Training}
We adopt the MT-DNN code base and adopt the exact optimization settings in \citet{liu2020mtmtdnn}. We use Adamax as our optimizer with a learning rate of $5\times 10^{-5}$ and a batch size of $32$. We train for a maximum number of epochs of $5$ with early stopping. A linear learning rate decay schedule with warm-up over $0.1$ was used. The dropout rate of all the task specific layers is set to be $0.1$, except $0.3$ for MNLI and $0.05$ for CoLa. We clipped the gradient norm within $1$. All the texts were tokenized using wordpieces, and were chopped to spans no longer than $512$ tokens. 

Worth mentioning, the task-specific super tickets used in Ticket Share are all selected during the case where a matched learning rate (i.e., $5\times 10^{-5}$) is used in single task fine-tuning. We empirically find that, rewinding the super tickets selected under a matched optimization settings usually outperforms those selected under a mismatched settings (i.e. using two different learning rates in single-task fine-tuning and rewinding/multi-task learning). This agrees with previous observation in literature of Lottery Ticket Hypothesis, which shows that unstructured winning tickets are not only related to its weight initialization, but also model optimization path.

\subsubsection{Multi-task Model Downstream Fine-tuning}
We follow the exact optimization setting as in Section~\ref{sec:st_exp_training} and in Section~\ref{app:st-training}, except we choose learning rate in $\{1\times10^{-5}, 2\times10^{-5}, 5\times10^{-5}, 1\times10^{-4}, 2\times10^{-4}\}$, and choose the dropout rate of all task specific layers in $\{0.05, 0.1, 0.2, 0.3\}$.

\subsection{Domain Adaptation Experiments}
\subsubsection{Data}
\label{app:da-data}

\noindent \textbf{SNLI.} is one of the most widely used entailment dataset for NLI. 

\noindent \textbf{SciTail} involves assessing whether a given premise entails a given hypothesis. In contrast to other entailment datasets, the hypotheses in SciTail is created from science questions. These sentences are linguistically challenging. The corresponding answer candidates and premises come from relevant web sentences. The lexical similarity of premise and hypothesis is often high, making SciTail particularly challenging.

Details of the SNLI and SciTail, including tasks, statistics, and evaluation metrics, are summarized in Table~\ref{tab:glue}.

\subsubsection{Training}
\label{app:da-training}
For single task model domain adaptation from MNLI to SNLI/SciTail, we follow the exact optimization setting as in Section~\ref{sec:st_exp_training} and in Section~\ref{app:st-training}, except we choose the learning rate in $\{5\times10^{-5}, 1\times10^{-4}, 5\times10^{-4}\}$.

\subsection{Sensitivity Analysis}
\label{app:ana-sensitivity}
\subsubsection{Randomness Analysis}
For single task experiments in Table~\ref{tb:seed_study}, we vary the random seeds only and keep all other hyper-parameters fixed. We present the standard deviation of the validation results over $5$ trails rewinding experiments. For multi-task downstream fine-tuning experiments,  we present the standard deviation of the validation results over $5$ trails, each result averaged over learning rates in $\{5\times 10^{-5}, 1\times 10^{-4}, 2\times 10^{-4}\}$. This is because the downstream fine-tuning performance is more sensitive to hyper-parameters.

\subsubsection{Hyper-parameter Analysis}
We further analyze the sensitivity of Ticket Share$\textsubscript{LARGE}$ model to changes in hyper-parameters in downstream fine-tuning in some GLUE tasks. We vary the learning rate in $\{5\times 10^{-5}, 1\times 10^{-4}, 2\times 10^{-4}\}$ and keep all other hyper-parameter fixed. Table~\ref{tb:lr_study} shows the standard deviation of the validation results over different learning rates, each result averaged over $5$ different random seeds. As can be seen, Task Share$\textsubscript{LARGE}$ exhibits stronger robustness to changes in learning rate in downstream fine-tuning.

\begin{table}[htb!]
\centering
\small
\begin{tabular}{@{\hskip3pt}l@{\hskip3pt}|C|C|C|C|C}
\toprule
& \textbf{RTE} & \textbf{MRPC} & \textbf{CoLA} & \textbf{STS-B} & \textbf{SST-2} \\ \midrule 
MT-DNN\textsubscript{LARGE}   & 1.26 & 0.86 & 1.05 & 0.42 & 0.26 \\ 
Ticket Share\textsubscript{LARGE}  & 0.44 & 0.58 & 0.61 & 0.36 & 0.25 \\ 
\bottomrule
\end{tabular}
\caption{Standard deviation of some tasks in GLUE (dev) over $3$ different learning rates.}
\label{tb:lr_study}
\end{table}

\subsection{Phase Transition on GLUE Tasks}
\label{app:phase-transition}
\begin{figure}[!htb]
    \centering
    \hspace{-0.3in}\includegraphics[width=0.9\linewidth]{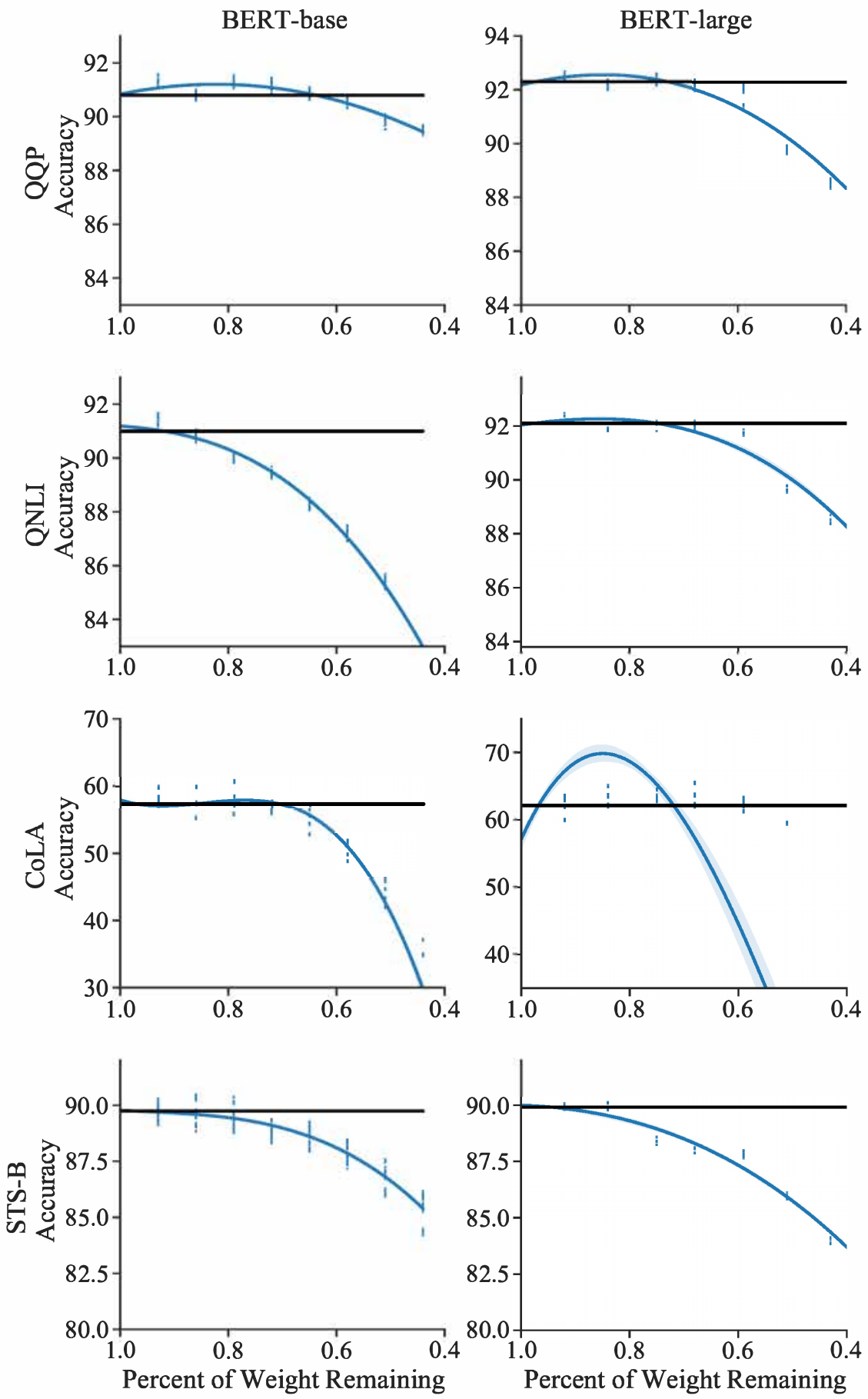}
	\caption{Single task fine-tuning evaluation results of the winning tickets on the GLUE development set under various sparsity levels.}
	\label{app:phase-transition-fig}
\end{figure}

Figure~\ref{app:phase-transition-fig} shows the phase transition plots on winning tickets on GLUE tasks absent from Figure~\ref{exp:phase_transtion}. All experimental settings conform to Figure~\ref{exp:phase_transtion}. 

\begin{table*}[htb]
	\begin{center}
		\begin{tabular}{@{ }l|l|c|c|c|c|c@{ }}
			\toprule 
			\bf Corpus &Task& \#Train & \#Dev & \#Test   & \#Label &Metrics\\ \midrule
			\multicolumn{6}{@{\hskip1pt}r@{\hskip1pt}}{Single-Sentence Classification (GLUE)} \\ \hline
			CoLA & Acceptability&8.5k & 1k & 1k & 2 & Matthews corr\\ \hline
			SST & Sentiment&67k & 872 & 1.8k & 2 & Accuracy\\ \midrule
			\multicolumn{6}{@{\hskip1pt}r@{\hskip1pt}}{Pairwise Text Classification (GLUE)} \\ \hline
			MNLI & NLI& 393k& 20k & 20k& 3 & Accuracy\\ \hline
            RTE & NLI &2.5k & 276 & 3k & 2 & Accuracy \\ \hline
			QQP & Paraphrase&364k & 40k & 391k& 2 & Accuracy/F1\\ \hline
            MRPC & Paraphrase &3.7k & 408 & 1.7k& 2&Accuracy/F1\\ \hline
			QNLI & QA/NLI& 108k &5.7k&5.7k&2& Accuracy\\ \midrule
			\multicolumn{6}{@{\hskip1pt}r@{\hskip1pt}}{Text Similarity (GLUE)} \\ \hline
			STS-B & Similarity &7k &1.5k& 1.4k &1 & Pearson/Spearman corr\\ \bottomrule
			\multicolumn{6}{@{\hskip1pt}r@{\hskip1pt}}{Pairwise Text Classification} \\ \hline
			SNLI & NLI& 549k &9.8k&9.8k&3& Accuracy\\ \hline
			SciTail & NLI& 23.5k &1.3k&2.1k&2& Accuracy\\ \bottomrule
		\end{tabular}
	\end{center}
	\caption{Summary of the GLUE benchmark, SNLI and SciTail.}
	\label{tab:glue}
\end{table*}